\documentclass[lettersize,journal]{IEEEtran}
\usepackage{amsmath,amsfonts}
\usepackage{algorithmic}
\usepackage{algorithm}
\usepackage{array}
\usepackage[caption=false,font=normalsize,labelfont=sf,textfont=sf]{subfig}
\usepackage{textcomp}
\usepackage{stfloats}
\usepackage{url}
\usepackage{verbatim}
\usepackage{graphicx}
\usepackage{cite}
\usepackage[dvipsnames]{xcolor}
\usepackage{lipsum}
\usepackage{amsmath,amssymb}
\usepackage{xcolor}
\usepackage{array}
\usepackage{multirow}
\usepackage{booktabs}

\title{Delving Deeper Into Astromorphic Transformers}

\author{Md Zesun Ahmed Mia,~\IEEEmembership{Student Member,~IEEE}, Malyaban Bal,
Abhronil Sengupta,~\IEEEmembership{Senior Member,~IEEE}
\thanks{MZA Mia, M. Bal and A. Sengupta are with the School of Electrical Engineering and Computer Science, The Pennsylvania State University, University Park, PA 16802, USA. E-mail: zesun.ahmed@psu.edu, mjb7906@psu.edu, 
sengupta@psu.edu.}}

\begin{document}
\maketitle

\begin{abstract}
Preliminary attempts at incorporating the critical role of astrocytes—cells that constitute more than 50\% of human brain cells—in brain-inspired neuromorphic computing remain in infancy. This paper seeks to delve deeper into various key aspects of neuron-synapse-astrocyte interactions to mimic self-attention mechanisms in Transformers. The cross-layer perspective explored in this work involves bioplausible modeling of Hebbian and presynaptic plasticities in neuron-astrocyte networks, incorporating effects of non-linearities and feedback along with algorithmic formulations to map the neuron-astrocyte computations to self-attention mechanism and evaluating the impact of incorporating bio-realistic effects from the machine learning application side. Our analysis on sentiment and image classification tasks (IMDB and CIFAR10 datasets) highlights the advantages of Astromorphic Transformers, offering improved accuracy and learning speed. Furthermore, the model demonstrates strong natural language generation capabilities on the WikiText-2 dataset, achieving better perplexity compared to conventional models, thus showcasing enhanced generalization and stability across diverse machine learning tasks.
\end{abstract}


\section{Introduction}
\IEEEPARstart{A}{strocytes}, a type of glial cell, play a critical role in brain function, encompassing various processes such as homeostasis, metabolism, and synaptic regulation \cite{perez2021role}. Astrocytes detect and regulate synaptic activity in the tripartite synapse through interactions with pre- and postsynaptic neurons. Investigating their impact on neural computation is currently an active research field in neuroscience and underscores the critical need to move beyond the neuro-synaptic perspective of current Artificial Intelligence (AI) systems. Recent experimental findings on neuron-astrocyte interactions and modulation have led to significant progress in computational neuroscience, enabling the development of models that incorporate neuron-astrocyte interactions within neural networks \cite{manninen2023analysis, wakida2020calcium}. Astrocytes have been found to modulate bursting in neural circuitry through the release of gliotransmitters, which have an impact on neuronal excitability and synaptic plasticity \cite{durkee2021astrocyte,lenk2020computational}. Astrocytes possess the ability to encode information through calcium signaling and regulate information processing, thereby actively engaging in neural computation at the tripartite synapse level. Additionally, astrocytes possess inherent capacity as memory components \cite{tsybina2022astrocytes,gordleeva2022situation} and plasticity regulators that are capable of facilitating local sequential learning \cite{gibbs2008astrocytic,alberini2018astrocyte}.

Preliminary attempts at incorporating the critical role of astrocytes in brain-inspired neuromorphic computing have started in earnest. Neuromorphic systems \cite{sengupta2019going} are bioplausible AI models that attempt to emulate neuronal and synaptic operations with a higher degree of bio-fidelity to tap into the huge energy efficiency of the brain. Such systems are temporal and event-driven, where computation occurs through transmission of sparse spiking events between neurons, thereby offering the potential of orders of magnitude power efficiency when implemented on neuromorphic platforms \cite{davies2021advancing}. However, astromorphic computing that encapsulates the features of astrocytic regulation in neuromorphic systems still remains in infancy, with works focusing primarily on the integration of astrocyte functionality with Spike-Timing-Dependent-Plasticity (STDP) mechanisms \cite{liu2018exploring,rastogi_self-repair_2021,han2022astromorphic, liu2017spanner}—an unsupervised learning framework that remains limited to shallow networks and simple tasks \cite{diehl2015unsupervised}. Other attempts also suffer from the pressing issue of scalability to complex tasks \cite{han2023astronet}.

This work is based on a recent study \cite{kozachkov2022building} that has shown promise at emulating the self-attention mechanism in transformers \cite{vaswani2017attention} using a simple two-layered network involving neuron-astrocyte interactions. The ability to design Astromorphic Transformers is therefore highly relevant in the current era of Large Language Models (LLMs), such as BERT \cite{devlin2018bert} and GPT \cite{radford2018improving}, paving the pathway for designing explainable transformers from a neuroscience perspective as well as addressing scaling challenges for utilizing the computational role of astrocytes in neuromorphic computing. However, as we illustrate in this work, prior work has neglected various key aspects of \textit{bi-directional astrocytic signalling, feedback, and non-linear temporal behavior}—all of which are key to addressing the fundamental mapping of self-attention mechanism to astrocyte-neuron interactions along with providing insights on whether such intrinsic non-linear dynamics and plasticity modulation enable better learning capabilities from the application side. The distinguishing feature of this work, therefore, lies in the \textbf{neuroscience-algorithm-application co-design} perspective seeking to answer the following questions: \textbf{(i) Theoretical neuroscience:} What aspects of astrocyte functionality should be included in neural architectures to better explain the self-attention mechanism in transformers? \textbf{(ii) Algorithm}: Can we map key algorithmic modules, like the relative positional information encoding of tokens, to astrocytic interactions by drawing qualitative inspiration from theoretical neuroscience? \textbf{(iii) Application:} What is the impact of incorporating such dynamic temporal bioplausible transformers from the machine learning application perspective?

\section{Related Works and Main Contributions}
Spiking neural networks (SNNs), inspired by biological neurons, have garnered attention for their ability to process temporal data efficiently through sparse, event-driven computation and communication \cite{pfeiffer2018deep,tavanaei2019deep}. Recent works have explored enhancing the computational efficiency of SNNs, including incorporation of advanced learning techniques \cite{zenke2018superspike}, improving network robustness through hybrid models \cite{mostafa2017supervised}, and addressing energy efficiency challenges \cite{yang2023sibols,yang2023snib}. While these advances contribute to energy-efficient and robust learning systems, they remain focused on spike-based computation and do not fully leverage neuron-astrocyte interactions, which is the core focus of our work.

Understanding the role of astrocytes in neuromorphic computing, especially from the algorithm perspective, is relatively nascent. While hardware implementation of astrocyte functionality in CMOS  \cite{ranjbar2015analog,karimi2018neuromorphic} as well as post-CMOS technologies \cite{garg2021emulation} have been explored, algorithm-level works exploiting the role of astrocytes and leveraging their computational role have largely remained limited to self-repair functionality of faulty neural network hardware \cite{wade_self-repair_2012, liu2018exploring,rastogi2021self,han2022astromorphic}. These works are mainly inspired by the fact that astrocytes, through their synaptic plasticity regulation and feedback mechanisms, enable neurons to fire at a baseline ideal frequency. This property can be utilized to mitigate neuromorphic hardware non-idealities with stuck-at-faults and drift effects \cite{han2022astromorphic}. However, such approaches are still limited to single-layered unsupervised networks where astrocyte functionality is integrated with STDP learning mechanisms. 
Recent studies have also examined the influence of astrocytes on learning \cite{alberini2018astrocyte}, working memory \cite{tsybina2022astrocytes}, energy minimization in neural networks \cite{fountain2022effect}, robotic locomotion control \cite{polykretis2020astrocyte}, structure learning \cite{han2023astronet}, among others. 

Building upon recent works \cite{kozachkov2022building} at emulating self-attention module in transformers through astrocyte-neuron interactions, we develop a computational framework that encapsulates key aspects of temporal non-linearities and feedback enabled by astrocytes and underscore the importance of incorporating such bioplausible effects through a comprehensive evaluation of Astromorphic Transformers on a benchmark evaluation set. Our work expands the computational capabilities of astrocyte-neuron network architectures equipped with plasticity, for the first time, to a broad category of sequential tasks.

\section{Methodology}
\subsection{Transformers}
Vanilla transformers \cite{vaswani2017attention} for classification tasks usually consist of an encoder block processing the provided \textit{key}$(K)$, \textit{query}$(Q)$, and \textit{value}$(V)$ generated from the input token embeddings $X$ (words in language tasks, pixels in vision tasks, etc.). Appendix A in the supplementary material illustrates the conventional configuration of a transformer, featuring a block referred to as ``Multi-Head Attention" that produces the contextual association among the provided inputs. The multi-head attention block in transformers uses a self-attention mechanism for multiple input heads, allowing the model to allocate varying importance to individual tokens, focusing on salient features for precise output in specific contexts.

The transformer implements auto-regressive attention by computing the dot product of \textit{key} and \textit{query} and then multiplying the resulting scalar with the corresponding \textit{value} derived from the input token. Implementing the softmax function (a type of distribution function) atop the dot product of \textit{key} and \textit{query} (to enforce attention over the tokens) increases the order of complexity, a predicament that can be alleviated by the use of linearized attention \cite{schlag2021linear}. Linearized attention exploits the distributive property of matrix multiplication, resulting in lower complexity for self-attention computation. More details can be found in Appendix B of the supplementary material. From the linear attention-based transformer (linear transformer) perspective \cite{katharopoulos2020transformers}, the transformer equations can be defined in the following form:
\begin{equation}\label{eq:1}
    \mathrm{SA} (X) = \frac{\phi(Q)(\phi(K)^TV)}{\phi(Q) [\sum_{t=1}^{t=N} \phi(k_t)]^T} \qquad \qquad\qquad \qquad 
\end{equation}
\begin{equation}\label{eq:Y}
    Y = \mathrm{LayerNorm} (\mathrm{SA}(X) + X) \qquad \qquad\qquad \qquad \;
\end{equation}
\begin{equation}\label{eq:Z}
    \mathrm{logits} = \mathrm{Softmax}(\mathrm{Linear}(\mathrm{LayerNorm} (\mathrm{FFN}(Y) + Y)))
\end{equation}
From Equations \ref{eq:Y} and \ref{eq:Z}, we can see that the output of self-attention is $Y$, which undergoes layer normalization after being augmented with residual connections from the input layer. The transformer determines the final output probabilities, denoted by logits. $\mathrm{FFN}$ stands for a Feed Forward Network and there is a $\phi(.)$ feature map that corresponds to a positive similarity function. When two such functions are multiplied, the normalized dot product maps to a distribution function \cite{chaudhari2021attentive} (\textit{i.e.}, softmax). According to \cite{katharopoulos2020transformers}, $\phi(.)$ can be defined by Equation \ref{eq:elu}. Here, the activation function includes $elu(x)$, which stands for exponential linear unit and is utilized for efficient and precise learning \cite{clevert2015fast}. 
\begin{equation}\label{eq:elu}
    \phi(x) = \mathrm{elu}(x) + 1 
\end{equation}
In the next section, we discuss the tripartite synapse model and aspects that will be necessary in constructing Astromorphic Transformers.

\subsection{Tripartite Synapse Dynamics}
\begin{figure}
\centering
\includegraphics[width=0.85\columnwidth]{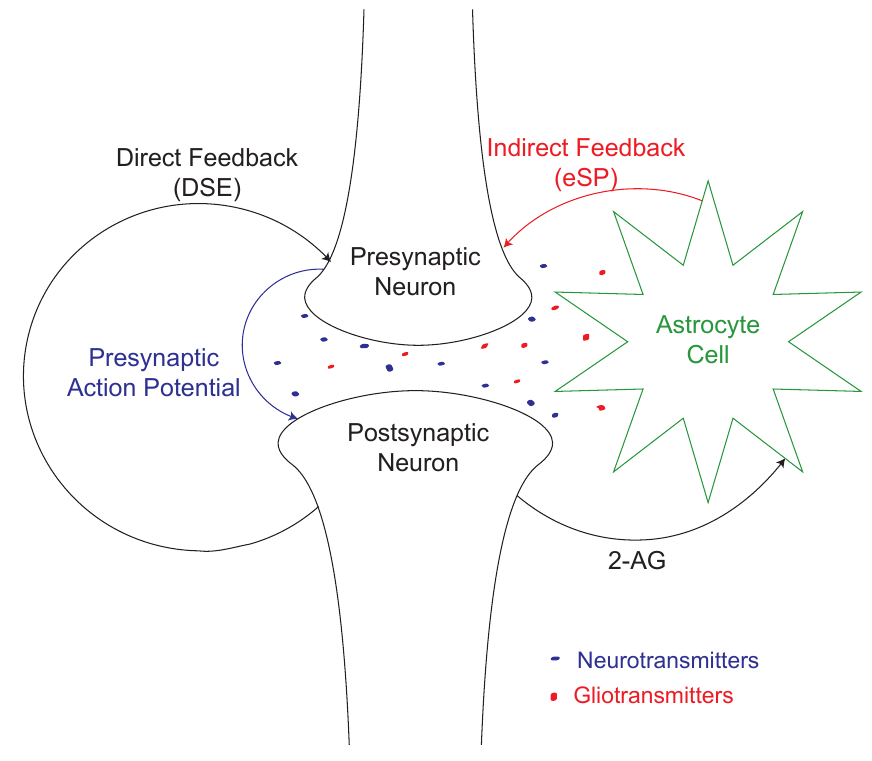} 
\caption{A model of synaptic communication in the brain. The tripartite synapse consists of presynaptic neurons, postsynaptic neurons, and astrocytes. Astrocytes detect neuronal activity and respond bidirectionally by emitting gliotransmitters, thereby modulating the intensity and duration of synaptic communication.}
\label{fig2:tripartite synapse}
\end{figure}

Astrocytes play a crucial role in the modulation of synaptic plasticity by releasing chemical signals that regulate the efficacy of synaptic connections. This phenomenon plays a role in modulating the strength and duration of synaptic connectivity and is crucial for maintaining the health of neurons. The depiction of the tripartite synapse in Fig. \ref{fig2:tripartite synapse}, which incorporates astrocytes, is a biologically plausible representation that embodies a multifaceted and evolving comprehension of synaptic signaling within the brain. 

\subsubsection{Astrocyte dynamics}
The tripartite synapse's dynamics are based on Li and Rinzel $Ca^{2+}$ dynamics \cite{li1994equations}, gatekeeper model \cite{volman2007astrocyte}, Nadkarni and Jung model \cite{nadkarni2004dressed, nadkarni2007modeling}, and bidirectional study on astrocytic glutamate binding to postsynaptic neurons (AN model) \cite{wade2011bidirectional}.
According to the biophysical models, it is observed that an action potential (AP) is generated by the presynaptic axon through the release of neurotransmitters. Neurotransmitters bind to postsynaptic dendrites, causing depolarization. This leads to the release of 2-Arachidonoylglycerol (2-AG) from postsynaptic neurons to type 1 Cannabinoid Receptors (CB1Rs) on the presynaptic terminal. This process is called Depolarization-induced Suppression of Excitation ($DSE$) and depresses the continuous release of neurotransmitters by inhibiting presynaptic neurons \cite{alger2002retrograde}. Neurotransmitter release also involves binding 2-AG to CB1Rs on an astrocyte, enveloping the synapse, and increasing Inositol 1,4,5-trisphosphate ($IP_3$) levels. $IP_3$ facilitates the release of calcium ions into the cytoplasm, thereby increasing calcium concentration. Three currents, $J_{channel}$, $J_{leak}$, and $J_{pump}$, are introduced to model the rate of change of calcium concentration inside the astrocyte \cite{li1994equations}. When calcium concentration crosses a threshold, astrocytes discharge glutamate, which binds to presynaptic group I metabotropic Glutamate Receptors (mGluRs), creating a process called eSP (Indirect Feedback) from the astrocyte to the presynaptic neuron \cite{navarrete2010endocannabinoids}. The comprehensive description of the computational model (Equation \ref{eq:ca-diff-eqn} introduces the basic equation of the model) can be found in the works of \cite{wade2012self,wade2011bidirectional, de2009glutamate}, which are outlined in Appendix D in the supplementary material.
\begin{equation}\label{eq:ca-diff-eqn}
    \frac{d(Ca^{2+})}{dt} = J_{channel} + J_{leak} - J_{pump}
\end{equation} 

\subsubsection{Neuron dynamics}
In this work, we consider the passive Leaky Integrate and Fire (LIF) dynamics to describe the neuron firing process, which is described in Equation \ref{eq:mem-potential}:
\begin{equation}\label{eq:mem-potential}
    \tau_m \frac{dv}{dt} = -v(t) + R_m\sum_{i=1}^{M} I_{total-syn}^i 
\end{equation}
where $\tau_m$ represents the membrane time constant, $v$ represents the membrane potential, and $R_m$ represents the membrane resistance. $I_{total-syn}^i$ denotes the total synaptic current at the $i^{th}$ synapse. $M$ is the number of synapses connecting to the corresponding neuron. Total synaptic current at the $i^{th}$ synapse can be defined as the summation of two currents, namely, Synaptic Current, $I_{syn}^i$ and Slow Inward Current, $SIC(t)$ in Equation \ref{eq:synaptic-current}:
\begin{equation}\label{eq:synaptic-current}
    I_{total-syn}^i = I_{syn}^i + SIC(t)
\end{equation}
$I_{syn}^i$ is the intrasynaptic current, which varies linearly with the amount of neurotransmitters present in that synapse at any time, $t$ ($I_{syn}^i = A_{se}y_i(t)$, where $A_{se}$ and $y_i(t)$ represent the absolute synaptic efficacy and the amount of neurotransmitter discharged by the synapse at a given time $t$, respectively). Thus, \textit{$I_{syn}$ depends on the direct feedback (corresponds to $DSE$)} as direct feedback is modulated by the amount of neurotransmitters between the pre- and postsynaptic neuron. On the other hand, the extrasynaptic \textit{Slow Inward Current $(SIC(t))$ is generated when astrocyte-released glutamate interacts with extrasynaptic NMDARs} (located in postsynaptic neurons) \cite{haydon2006astrocyte}. To elaborate, the activation of astrocytes leads to the release of glutamate from these cells. Upon its release, glutamate interacts with NMDARs located on the postsynaptic neuron, resulting in the emergence of a distinct form of current known as Slow Inward Current $(SIC(t))$ \cite{wade2011bidirectional}. The details of the mathematical modeling of $SIC(t)$ can be found in \cite{wade2011bidirectional}.
Next, we develop a network architectural framework to mimic the self-attention mechanism in transformers using the astrocyte and neuron dynamics described herein and evaluate the impact of inherent non-linearities and feedback.

\begin{figure}
\centering
\includegraphics[width=0.77\columnwidth]{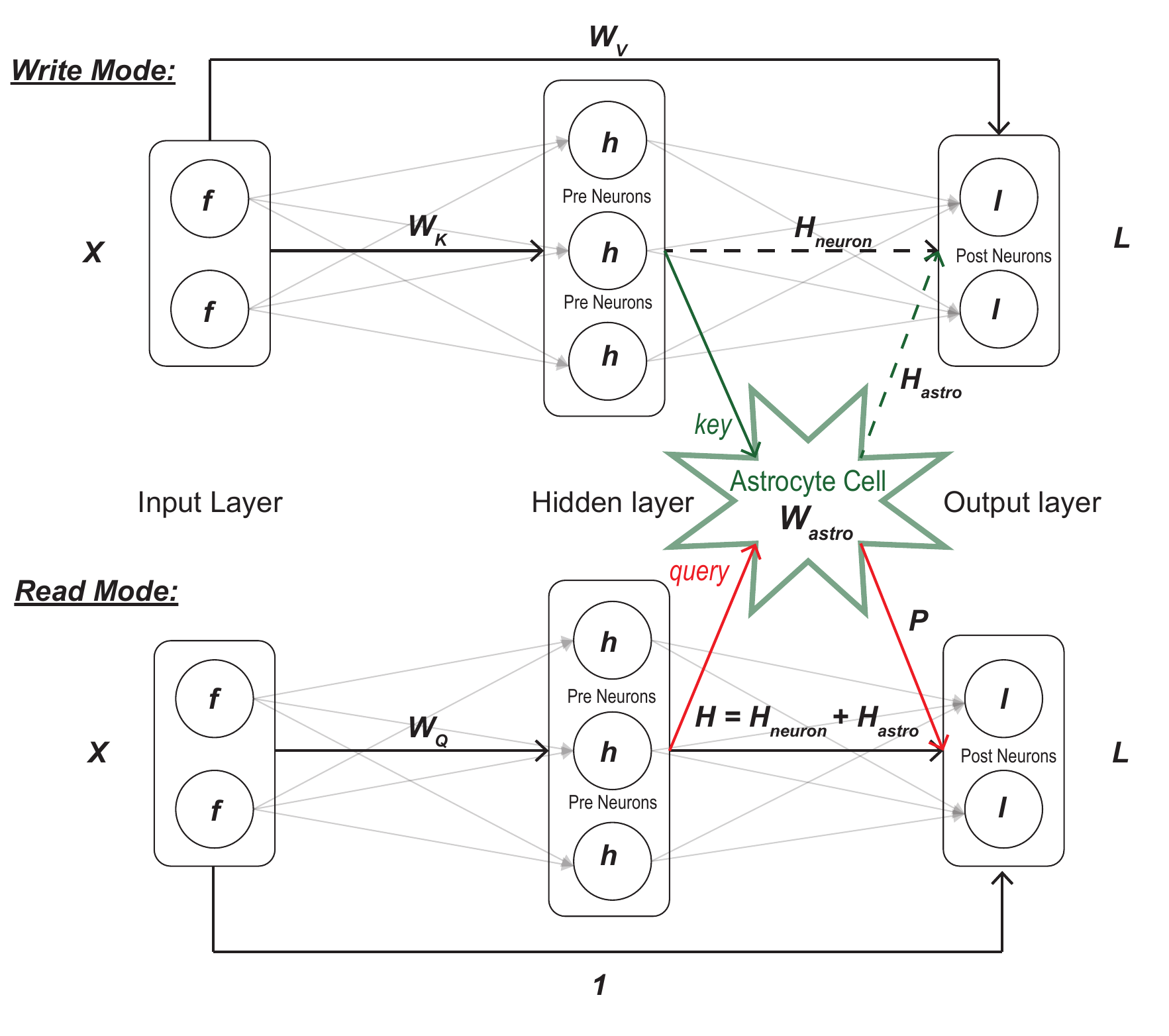} 
\caption{The neural network architecture showing the three layers. As tokens are presented to the network as a $d$ dimensional vector, there are $d$ neurons in both the input and output layers. The hidden layer has $m$ neurons. Solid lines indicate active operations; therefore, $H_{neuron}$ and $H_{astro}$ are learned during write mode but utilized during read mode.} 
\label{fig3:neural-network}
\end{figure}

\subsection{Neural Network Emulating Self-Attention }
Self-attention is a fundamental component of all transformer models. Based on prior works of mimicking self-attention by astrocytes \cite{kozachkov2022building}, we adopt a network topology as shown in Fig. \ref{fig3:neural-network}. The neural network architecture comprises three distinct layers, namely input layer, hidden layer, and output layer. There exist two distinct operational modes for the aforementioned neural network: (i) write mode and (ii) read mode. Upon presentation of the tokens ($x_t \in R^{1\times d}$: which may include word embeddings and image embeddings) to the input layer, the write mode is initially employed, followed by the read mode to generate the outputs. In write mode, we encode information into the network by means of \textit{keys} and \textit{values}, whereas, in read mode, we retrieve the information by providing \textit{queries}. 

During the write mode, the input layer sequentially transmits the \textit{keys} ($k_t = x_tW_K ; \; W_K \in R^{d \times m}$) for each token to the hidden layer and the \textit{values} ($v_t = x_tW_V ; \; W_V \in R^{d \times d}$) to the output layer. The hidden layer neurons introduce a non-linearity within the system by activating a feature map, denoted as $\phi(.)$, on the transmitted \textit{keys}. The use of $\phi(.)$ reduces the computational complexity and memory usage inspired from the linearized attention mechanism. More information on the reduction of the order of complexity is discussed in Appendix B (with Fig. S2 illustrating the idea) in the supplementary material. A presynaptic neuron in the hidden layer ($ h_t \in R^{1 \times m} $) is intricately connected with an astrocyte and an output layer ($ l_t \in R^{1 \times d} $) neuron, forming a tripartite synapse. Similar to the mechanism by which the presynaptic neuron stimulates the astrocyte through the release of neurotransmitters, in this context, the astrocyte is being activated through the transmission of \textit{keys} ($ k_t \in R^{1 \times m} $) from neurons in the hidden layer. The write mode can be mathematically expressed as in Equation \ref{eq:write-mode}:
\begin{equation}\label{eq:write-mode}
    \begin{aligned}
      f_t = x_t \qquad \qquad \qquad  \; \; \;\\
      h_t = \phi(f_tW_K) = \phi(k_t)   \\
      l_t = f_tW_V = v_t\qquad \quad 
    \end{aligned}
\end{equation}

In order to establish the read mode equations, it is necessary to first formulate the neural plasticities that occur among the components of the tripartite synapse. Moreover, our work aims at exploring strong bioplausible correlations in modeling such plasticities, which ultimately influence our neuroscience-algorithm-application co-design analysis. In a subsequent section, we will examine the read mode, which is responsible for retrieving the encoded information.

\subsection{Mathematical Formulation of Plasticity}
\subsubsection{Hebbian Plasticity}
Hebbian plasticity is characterized by the sequential adjustment of synaptic weights in accordance with the interconnection pattern between two neurons (pre- and postsynaptic neurons) at the synapse. Previous studies \cite{kozachkov2022building} have exclusively focused on the plasticity occurring between two neurons, neglecting non-linearities and the plasticity that may occur between astrocytes and postsynaptic neurons, which is also involved in the learning process. 

\textbf{Contribution I:} In the context of Hebbian plasticity, we underscore the importance of the addition of astrocyte-postsynaptic neuron connected Hebbian-inspired learning along with inherent non-linearities. To elaborate, the synaptic weights ($H_{neuron}$) that establish connections between the presynaptic and postsynaptic neurons are acquired through Hebbian plasticity during the write mode. This learning occurs due to the connection between pre- and postsynaptic neurons, which can be mapped to the synaptic current ($I_{syn}$) from the neuroscience perspective. As discussed previously, $I_{syn}$ is modulated by \textit{DSE}, so $H_{neuron}$ is inspired from the \textit{direct feedback} in the tripartite synapse. This leads to Hebbian plasticity between the hidden layer (presynaptic neuron) and the output layer (postsynaptic neuron). Hidden neuron outputs are denoted as $h_t = \phi(k_t)$, whereas the output (last) layer outputs are denoted as $l_t = v_t$ in the write mode (refer to Equation \ref{eq:write-mode}). $H_{neuron}$ weights are learned via Equation \ref{eq:H-neuron}, where $m$ acts as a scaling factor affecting the learning speed.
\begin{equation} \label{eq:H-neuron}
    H_{neuron,t} = H_{neuron,t-1} + \frac{1}{m}h_{t}^T l_{t}
\end{equation}
The astrocyte functions as the third constituent in the tripartite synapse, enabling bidirectional communication with the synaptic neurons. Hence, we propose that the connection between the astrocyte and the postsynaptic neuron also be represented in this formulation through a learnable weight matrix ($H_{astro}$), in accordance with the Hebbian rule. Equation \ref{eq:H-astro} shows the learning mechanism of these weights in which the presynaptic neuron output ($h_t$) is replaced by the astrocytic activity, represented by $w_{astro}$, as this connection incorporates the astrocyte and the postsynaptic neuron. The formulation of $w_{astro}$ will be considered in the succeeding text.
\begin{equation} \label{eq:H-astro}
    H_{astro,t} = H_{astro,t-1} + \frac{1}{m} w_{astro} l_{t}
\end{equation}
The weight, denoted as $H_{astro}$, is inspired from the extrasynaptic current $SIC$ (refer to the tripartite synapse dynamics section). According to Equation \ref{eq:synaptic-current}, $SIC$ is added to $I_{syn}$ to generate the total synaptic current. Consequently, the two distinct learnable weights, denoted as $H_{neuron}$ and $H_{astro}$, are combined to produce the Hebbian weight $H = H_{neuron} + H_{astro}$.

Astrocytes have a significant impact on the Hebbian weight due to their inherent non-linearity. A recent study by \cite{han2023astronet} introduced an AstroNet model that optimizes neural network connections through temporal and global regulation mechanisms. The authors found that astrocytes introduce non-linear effects, which compound the pre-existing non-linearity of neurons. The concept of Hebbian plasticity is associated with temporal and global regulation, where the current weight is influenced by past neuron weights and astrocytic weights. To map the effect of non-linearity, we consider that the summation of neuron and astrocyte Hebbian weights undergoes non-linear activation, ultimately generating the Hebbian weight ($H$). As such, we propose the integration of a sigmoid ($\sigma$) non-linearity into the learning process. Thus, the weight $H$ exhibits the collective Hebbian plasticity of a tripartite synapse, as elucidated by Equation \ref{eq:HP}, where $K$ and $V$ correspond to \textit{keys} and \textit{values} of all the $N$ tokens in a single sequence (in matrix form: $K \in R^{N \times m}$, $V \in R^{N \times d}$). Assuming initial $H_{neuron,0}$ and $H_{astro,0}$ from Equations \ref{eq:H-neuron} and \ref{eq:H-astro} are both zero, we can define the Hebbian weight in its matrix form ($H \in R^{m \times d} $) as:
\begin{equation}\label{eq:HP}
\begin{aligned}
    H = \sigma(H_{neuron} + H_{astro})  \qquad  \qquad \qquad  \; \; \: \\
    H = \sigma( \frac{1}{m}(\phi(K)^T V + W_{astro} V ))   \qquad \quad \; \;  \\
\end{aligned}
\end{equation}

Next, we consider formulation of the astrocytic activity parameter $W_{astro}$ (Note, $w_{astro}$ in Equation \ref{eq:H-astro} is the per-token representation and $W_{astro}$ in Equation \ref{eq:HP} is the matrix representation). We note that astrocytic $Ca^{2+}$ dynamics is a relatively slower temporal process than the streaming rate of the input tokens. Therefore, it is expected that the astrocytic activity should encode relative information among successive tokens. In this context, we derive inspiration from algorithmic formulations of relative positional information of tokens, popularly used in Vision transformers \cite{shaw2018self, ramachandran2019stand}, and map it to the activity parameter $W_{astro}$. The astrocytic activity parameter is therefore defined to use the edge of the tokens to capture relative position differences between input elements \cite{shaw2018self}. The edge of the tokens considers the pairwise relationships between input elements in the sense that the input is modeled as a labeled, directed, fully connected graph. 
To model astrocytic activity, we first define the edges between consecutive input tokens $i$ and $j$ as \( a_{ij} \) with the help of matrix $P \in R^{N \times N}$, containing the distance-based relative positional information between tokens, and $D \in R^{m \times m}$ which is a transformed version of $P$, achieved through a matrix basis transformation $D = MPM^T$. In this transformation, $M \in R^{m \times N}$ is a learnable matrix that projects the original $N$-dimensional space of positional relationships into an $m$-dimensional space. This transformation allows the model to learn an optimal representation of positional information, potentially reducing dimensionality if $m < N$, and capturing complex relationships between token positions. Mathematically, this transformation is a linear transformation, specifically a quadratic form, suitable for encoding pairwise interactions. Conceptually, this can be viewed as a learned change of basis, where $M$ defines a new coordinate system that optimizes the representation of positional information for the task. The resulting matrix $D$ encodes the spatial relative distances among tokens, represented as neuronal activity received by the astrocyte. Based on $D$, the edges are formulated as: $a_{ij} = W_{rel} D$, where $W_{rel} \in R^{N \times m}$ is a randomly initialized learnable weight matrix. To further enhance the efficiency of our model, we set $W_{rel} = M^T$. This design choice offers several advantages: First, it reduces the number of learnable parameters, promoting more efficient training and mitigating the risk of overfitting, especially in data-limited scenarios. Second, it encourages the learning of a more compact and generalizable representation, as the transformation $M$ now serves the dual purpose of both transforming the positional information ($P$ to $D$) and weighting the transformed information for edge computation ($a_{ij} = M^T D$). Finally, this constraint acts as a form of regularization, guiding the model to learn more robust and meaningful features by ensuring that the learned transformation is useful for both representing and processing positional relationships. The parameter, $m$, representing the number of neurons in the hidden layer, is crucial as it dictates the dimensionality of the learned representation. Its value is not fixed but rather can be adjusted based on the complexity of the dataset, allowing the model to adapt its capacity. Datasets with higher complexity may benefit from a larger $m$, providing the model with more representational power, while simpler datasets might learn effectively with a smaller $m$, promoting efficiency and potentially preventing overfitting.
Building on the definition of edges, the astrocytic activity parameter \( W_{astro} \in R^{m \times N} \) (\(w_{astro} \in R^{m \times 1}\) for per token form) is defined as:
\begin{equation}\label{eq:edge-info}
    W_{astro} = \phi(a_{ij})^T  \\ 
\end{equation}
where \( \phi \) introduces a non-linearity to capture the temporal dynamics of astrocytes. 
While \( W_{rel} \) is learned over time, the positional information provided by \( D \) ensures that the model maintains awareness of relative distances among tokens, allowing \( W_{astro} \) to adapt dynamically to the input.
Consequently, $H_{astro}\in R^{m \times d}$ can be considered as a representation of acquired positional knowledge derived from the input tokens.

\subsubsection{Presynaptic Plasticity}
The plasticity resulting from the induced calcium response inside the astrocyte due to bidirectional communication between the astrocyte and presynaptic neurons (hidden layer) can be defined as presynaptic plasticity. As tokens are sequentially presented to the hidden layer during the write mode, \textit{keys}: $\phi(k_t)$ are generated, which can be mapped to an increasing number of released neurotransmitters (implying increment in $J_{channel}$ as per the computational neuroscience model), thereby resulting in elevation of the calcium concentration within the astrocyte (evident from Equation \ref{eq:ca-diff-eqn}). Thus, the presynaptic plasticity parameter $g$ can be mapped to calcium concentration in astrocytes modulated by indirect feedback ($eSP$), and can be expressed as the aggregate of the inputs ($h_t = \phi(k_t)$) received by the presynaptic neuron. Equation \ref{eq:PSP} defines the formulation of the presynaptic plasticity parameter ($g_t, g \in R^{1 \times m}$). Assuming the initial calcium concentration to be zero ($g_0 = 0$), $g$ can be expressed as:
\begin{equation}\label{eq:PSP}
\begin{aligned}
    g_t = g_{t-1} + h_t = g_{t-1} + \phi(k_t) \quad   (\mathrm{pertoken \; form})\\
    g = \sum_{t = 1}^{N} {\phi(k_t)} \: \; \; \quad \qquad\qquad \qquad\quad  \;  (\mathrm{matrix \; form})
\end{aligned}
\end{equation}

\begin{figure} [t]
\centering
\includegraphics[width=0.68\columnwidth]{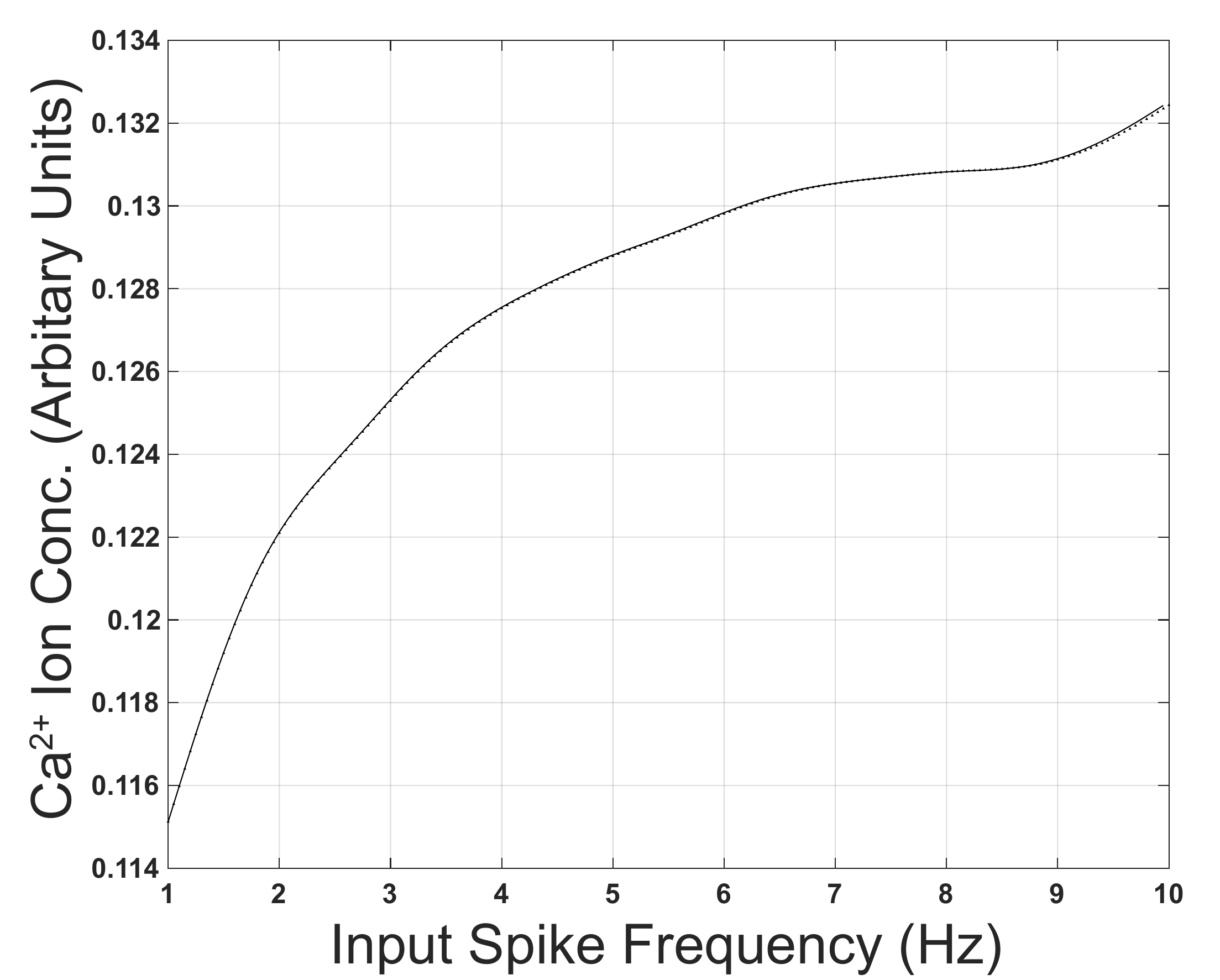} 
\caption{As the input spike frequency in the presynaptic neurons increases, the rate of increase in calcium concentration slows down, thereby exhibiting a non-linear relationship between $Ca^{2+}$ ion concentration and input spike rate.}
\label{fig5:ca-response}
\end{figure}

\textbf{Contribution II:} Prior work \cite{kozachkov2022building} at modelling pre-synaptic plasticity has not considered the intrinsic non-linear $Ca^{2+}$ dynamics of the astrocytes in response to increasing firing rate of presynaptic neurons. Due to the indirect feedback (\textit{eSP}), the rate of increase in the release rate of neurotransmitters gradually reduces in response to an increase in presynaptic neuron firing over time. 
Hence, a non-linear relationship exists between the calcium concentration and the frequency of spikes received from the presynaptic neuron. We substantiate this observation by simulating the AN model \cite{wade2012self, wade2011bidirectional}. The associated non-linear variation is depicted in Fig. \ref{fig5:ca-response} which leads us to modify the presynaptic plasticity parameter $g$ as,
\begin{equation}\label{eq:nonlinearity}
    g = (\sum_{t = 1}^{N} {\phi(k_t)})^\alpha
\end{equation}
where the exponent $\alpha$ encodes the degree of non-linearity. From our computational modeling, we use a value of $\alpha = 0.25$. The subsequent section elaborates on the read mode operation, which involves the process of retrieving the information encoded during the write mode to implement the self-attention mechanism incorporating our key contributions towards more bioplausible formulations of Hebbian and presynaptic plasticity.

\subsection{Realization of Self-Attention}
The self-attention mechanism is realized when the encoded information is accessed during the read mode to obtain the activations of the output layer neurons. The hidden layer neurons receive \textit{queries} ($q_t = x_tW_Q$) from the input layer and access the information that was previously encoded within the astrocyte. This information is stored via both the Hebbian weight ($H$) and the presynaptic plasticity parameter ($g$) during write mode. However, during read mode, $H$ is already learned and not modulated by the transmitted \textit{queries} from the presynaptic neurons, as it is directly connected to the postsynaptic neurons (output layer). 
As $g$ is the presynaptic plasticity parameter, it has no direct connection to the postsynaptic neuron. Therefore, the information encoded in $g$ has to be retrieved by the \textit{queries} transmitted from the hidden layer during read mode by inducing a calcium response from the astrocytes. 
In order to read the calcium response ($c_t \in R^{1 \times 1} $) of the astrocyte, the \textit{queries} ($h_t$) are scaled by the corresponding presynaptic plasticity parameters ($g_t$) across all the $m$ hidden neurons and then summed together. This is depicted in Equation \ref{eq:cal-resp}, where the matrix form of the calcium response is defined by $C \in R^{N \times 1} $.
\begin{equation}\label{eq:cal-resp}
\begin{aligned}
    c_t = \sum_{m}  g_t h_t  \quad   (\mathrm{pertoken \; form})\\
    C =  h g^T \: \; \quad \qquad  (\mathrm{matrix \; form})
\end{aligned}
\end{equation}

Let us define the weight modulating the connection between the astrocyte and the output layer as $P$. According to studies regarding Tumor Necrosis Factor - $\alpha$ ($TNF - \alpha$), astrocytes possess the ability to boost synaptic weights in reaction to low neural activity and diminish weights in reaction to high neural activity \cite{kozachkov2022building} by means of calcium response (neural activity is proportional to released neurotransmitters). Consequently, it is possible to infer an inverse correlation between $P$ and the calcium response ($C$). Simplifying, this relation (\(P=\frac{1}{C}\)) is given by Equation \ref{eq:P-weight}. To avoid the computational cost of a full matrix inversion, we compute the reciprocal calcium response element-wise by inverting only the non-zero elements of the diagonal matrix using a mask. This ensures numerical stability and computational efficiency for large-scale sequences. 
\begin{equation}\label{eq:P-weight}
    P = \frac{1}{\phi(Q) [(\sum_{t = 1}^{N} {\phi(k_t)})^\alpha]^T }
\end{equation}

Given that the presynaptic plasticity modulated weight ($P$) is ascertained, we can define the network equations for the read mode. As seen from Fig. \ref{fig3:neural-network}, the hidden layer and the output layer are connected by the Hebbian weight $H$, and the astrocyte is imposing another learned weight (from calcium response) denoted by $P$ onto the existing weight $H$. Thus the weights are implemented by the Hadamard (element-wise: $\odot$) multiplication of $H$ and $P$ onto the output layer. Furthermore, the output layer receives the input ($F$) directly from the input layer modulated by unity weight to implement the residual connection present in the self-attention module of transformers. Considering the network architecture, the read mode network equations can be defined in Equation \ref{eq:read-mode}. It is important to highlight here that during the read mode, the decoding process occurs in a singular timestep, as opposed to the sequential nature of the write mode. Hence, all the parameters for input, hidden, and output layers are represented in their matrix format (e.g., \(X\in R^{N \times d}\), \( h\in R^{N \times m} \), \(L\in R^{N \times d}\) ).
\begin{equation}\label{eq:read-mode}
    \begin{aligned}
      F = X \qquad \qquad \qquad \qquad \qquad \qquad \qquad \qquad  \; \: \\
      h = \phi(FW_Q) = \phi(Q) \qquad \qquad \qquad \qquad \qquad \;  \\
      L = h (H \odot P) + F = \phi(Q)(H\odot P) + X  \qquad  
    \end{aligned}
\end{equation}
In Equation \ref{eq:read-mode}, all the parameters are in matrix mode, where the dimensions are defined as follows: $W_Q \in R^{d \times m}$ and $Q \in R^{N \times m}$ (\textit{queries}). $L$ represents the generated output from the last layer. The self-attention algorithm is defined when the network completes a write and read mode, as discussed next.

By substituting the values of $H$ and $P$ from Equations \ref{eq:HP} and \ref{eq:P-weight} into Equation \ref{eq:read-mode}, the resulting Equation \ref{eq:astro-transformer-matrix} can be obtained after the write and read mode:
\begin{equation}\label{eq:astro-transformer-matrix}
    \begin{aligned}
    L = \frac{\phi(Q)(\sigma(\phi(K)^TV + W_{astro}V))}{m \times \phi(Q) [(\sum_{t = 1}^{N} {\phi(k_t)})^\alpha ]^T } + X \quad \; \;  \\
    Y = \mathrm{Layernorm}(L) \qquad \qquad \qquad \qquad \qquad \; \;  \\
    \mathrm{logits} = \mathrm{Softmax}(\mathrm{Linear}(\mathrm{FFN}(Y)+Y)) \quad \; \;
    \end{aligned}
\end{equation}
In Equation \ref{eq:astro-transformer-matrix}, $L$ implements an identical operation as the transformer self-attention from Equation \ref{eq:1} (along with the inclusion of an extra residual connection from the input layer). $Y$ is the layer-normalized version of $L$ and is further feedforwarded through fully connected linear layers to finally obtain the output probabilities in the form of logits. Thus, with this implementation, the Astromorphic Transformer is formulated. The detailed implementation of this process, including neuron-astrocyte interactions and synaptic modulation, is provided in Appendix E in the supplementary material, where the pseudo-code outlines the sequence of operations governing synaptic plasticity and calcium-mediated attention adjustments in the Astromorphic Transformer.

\begin{table*}[h!]
    \small
    \caption{Performance comparison of our proposed astromorphic attention-based transformer architecture against other benchmarks with varying degrees of bioplausibility on the IMDB, CIFAR10 and Wikitext-2 datasets.}
    \centering
    \begin{tabular}{ m{4.5em}||m{15.5em}m{19em}m{4em} m{5em} }  %{|c|c|c|c|c|c|} %
     \hline
     Task & Model & Implementation & Bio-plausibility & Accuracy(\%)/ Perplexity \\  
     \hline
     \hline

     \multirow{15}{4.5em}{Sentiment Classification (IMDB Dataset)} &Memristor SNN \cite{huang2023text}&Directly trained in SNN domain&Yes& $84.86\%$\\ %[-0.5ex]

     %\cmidrule{3-5}

     &Memristor SNN \cite{huang2023text}&Converted from ANN&Yes& $85.88\%$ \\

     %\cmidrule{2-5}

     &ANN \cite{huang2023text}&No spiking activity&No& $86.02\%$ \\

     %\cmidrule{2-5}

     &Spike-transformer \cite{mueller2021spiking}& Rate-coded approach using spiking neurons& Yes &$86.36\%$ \\

     %\cmidrule{2-5}

     &Recurrent SNN \cite{wang2022recurrent}& Presynaptic current-based backpropagation&Yes& $86.82\%$\\

     %\cmidrule{2-5}

     &HRNaEMSA biGRU \cite{leng2021using} &Implements multi-headed attention&No& $87.3\%$ \\

     %\cmidrule{2-2}

     %\cmidrule{5-5}

     &HRNaEMSA biLSTM \cite{leng2021using} &Implements multi-headed attention&No&$87.4\%$ \\

     %\cmidrule{2-5}

     &CoRNN \cite{rusch2020coupled}&Coupled oscillators modeled as neurons&No&$87.4\%$\\

     %\cmidrule{2-5}

     & LSTM \cite{aswani2021performance} & Memristor crossbars mimicking NN & No & $87.48\%$ \\ 

     %\cmidrule{2-5}

     & SNN \cite{agrawal2021impulse} &  Compute-in-memory macro using SNN & Yes & $88.15\%$ \\

     %\cmidrule{2-5}

     &UniCoRNN \cite{rusch2021unicornn}&Solves vanishing gradient problem in BPTT&No& $88.38\%$\\

     %\cmidrule{2-5}

     &Vanilla Transformer&Softmax attention without bioplausibilities&No&$88.9\pm0.1\%$\\

      &Linearized Transformer inspired from astrocytes (Iso-architecture) \cite{kozachkov2022building}&Linear attention with no astrocytic non-linearity or relative positional encoding  &Yes&$88.4\pm0.1\%$ \\[1ex]

     \cmidrule{2-5}

     &\textbf{Astromorphic Transformer (Our Model)}&Implements astromorphic attention with non-linearity and relative positional information&Yes& $\mathbf{88.7\pm0.2\%}$\\[2ex]

     \hline

     \multirow{11}{4.5em}{Image Classification (CIFAR10 Dataset)} &BioLeaF \cite{yang2021bioleaf} &SNNs trained by BP-based rules&Yes&$86.88\%$  \\ 

     %\cmidrule{2-5}

     &STBP in CIFARNET \cite{zhou2023spikingformer}&Spatio-temporal backprop &No&$89.83\%$\\

     %\cmidrule{2-5}

     &VGG16-SNN \cite{srinivasan2020training}&ANN-SNN conversion&Yes&$91.55\%$\\

     %\cmidrule{2-5}

     &MS-ResNet \cite{hu2021advancing}&SNN oriented ResNet-110 architecture&No&$91.72\%$\\

     %\cmidrule{2-5}

     &ANN \cite{zhou2022spikformer}&ResNet-19 (12.63M parameters)&No&$94.97\%$\\

     %\cmidrule{2-5}

     &ANN \cite{zhou2022spikformer}&Transformer-4-384 (9.32M parameters)&No&$96.73\%$\\

     %\cmidrule{2-5}

     &Spikformer \cite{zhou2022spikformer}&Spikformer-4-384 (9.32M parameters)&Yes&$95.19\%$\\

     %\cmidrule{2-5}

     &Spikingformer \cite{zhou2023spikingformer}&Spikingformer-4-384 (9.32M parameters)&Yes&$95.61\%$\\

     %\cmidrule{2-5}

     &Vanilla Transformer&Softmax attention without bioplausibilities&No&$97.4\pm0.1\%$\\

      &Linearized Transformer inspired from astrocytes (Iso-architecture)\cite{kozachkov2022building}&Linear attention with no astrocytic non-linearity or relative positional encoding &Yes&$96.8\pm0.2\%$ \\[1ex]

     \cmidrule{2-5}
     &\textbf{Astromorphic Transformer (Our Model)}&Implements astromorphic attention with non-linearity and relative positional information&Yes&$\mathbf{97.0\pm 0.2\%}$\\[2ex]

     \hline
     \multirow{9}{4.5em}{Language Modeling (Wikitext-2 Dataset)}
     & LSTM \cite{merity2016pointer}& Traditional RNN-based language model& No& $65.9$\\
     & AWD-LSTM \cite{takase2018direct}& Standard LSTM with dropout, regularization& No& $58.8$\\

     & AWD-LSTM-MoS \cite{yang2017breaking}& Improved LSTM for language modeling& No& $63.88$\\

     & AWD-FWM \cite{schlag2020learning}& Fast weight memory-based language model& No& $61.65$\\

     &Vanilla Transformer&Softmax attention without bioplausibilities&No&$73.4$\\

     & Linearized Transformer inspired from astrocytes (Iso-architecture) \cite{kozachkov2022building}& Linear attention with no astrocytic non-linearity or relative positional encoding \textbf{(does not converge due to gradient explosion)} & Yes& $123.3$ \\

     \cmidrule{2-5}

     & \textbf{Astromorphic Transformer (Our Model)}& Implements astromorphic attention with non-linearity and relative positional information& Yes&$ \mathbf{33.8}$\\[2ex]

     \hline
    \end{tabular}
    \label{tab:accuracy-comparison}
\end{table*}

\section{Results}
Our Astromorphic Transformer is trained and tested on three different tasks: sentiment classification on the IMDB dataset \cite{maas-EtAl:2011:ACL-HLT2011}, image classification on the CIFAR10 dataset \cite{krizhevsky2009learning}, and language modeling on the Wikitext-2 dataset \cite{merity2016pointer}. 
To date, there has not been any language or vision model that incorporates transformer architecture in a bioplausible manner derived from neuron-astrocyte interactions. We report the following performance for each task:

\begin{itemize}
    \item $88.7\pm0.2\%$ for sentiment classification on the IMDB dataset, utilizing the non-linearity and relative positional information in the model.
    \item $97.0\pm0.2\%$ accuracy for image classification using an astromorphic vision transformer on the CIFAR10 dataset.
    \item A perplexity of $33.8$ for language modeling on the Wikitext-2 dataset using one decoder layer with 6-headed self-attention.
\end{itemize}

\subsection{Hardware Specifications \& Dataset Details}
The Astromorphic Transformer has been executed on a GPU-based platform running PyTorch. The hardware setup consists of a $12$-core Intel(R) Xeon(R) Platinum $8167M$ CPU @ 2.00GHz, with an NVIDIA® Tesla® V100 GPU (16 GB memory).

The IMDB dataset contains $50,000$ movie reviews labeled as positive or negative, with each review spanning between 250 and 500 words. The CIFAR10 dataset comprises $60,000$ color images categorized into 10 distinct classes, with each image having dimensions of $32 \times 32$ pixels. The Wikitext-2 dataset is a widely used benchmark for language modeling and natural language generation tasks, containing over two million words with carefully preserved capitalization, punctuation, and rare terms.

\subsection{Network Architecture}

For the sentiment classification task on the IMDB dataset, we employ a transformer with an encoder-based architecture, using one encoder layer with 4-headed self-attention. A dropout layer is applied after the residual connection before layer normalization.

For the image classification task on the CIFAR10 dataset, we implement the CoAtNet transformer architecture, combining convolutional neural networks (CNNs) with self-attention. The architecture employs a $3 \times 3$ convolution before creating patches, followed by an 8-layered encoder stack with 8-headed self-attention. A patch size of $2$ is used on the image inputs. In addition to the original CoAtNet architecture, we add non-linearity and average pooling after the encoder layers.

For the language modeling task on the Wikitext-2 dataset, we use a decoder-based transformer architecture with one decoder layer and 6-headed self-attention. The embeddings for this task are generated using the GPT-2 tokenizer, and they are passed through an astromorphic decoder block with layer normalization and dropout before the final output. 

For all tasks, we use AdamW optimizer and Cross Entropy Loss for training. For tokenization, the GloVe pretrained word embeddings \cite{pennington2014glove} are used with 840B tokens, 2.2M vocab (cased), and 300D vectors for the IMDB dataset, and for CIFAR10 dataset, pixels are converted to sequences by using a patch size of 2, while GPT-2 tokenizers are used for the Wikitext-2 dataset. All hyperparameters for these architectures are detailed in Appendix C.

\subsection{Performance Evaluation and Comparative Analysis}
Table \ref{tab:accuracy-comparison} compares the performance of our Astromorphic Transformer against both bioplausible neuromorphic models and non-bioplausible models on the IMDB, CIFAR10, and Wikitext-2 datasets. We chose not to compare our results against pretrained transformer models like BERT and GPT, as these architectures leverage pretraining on large datasets and use non-bioplausible self-attention mechanisms with highly customized learnable embeddings. In contrast, our model is trained from scratch, making it simpler to implement the astromorphic attention in the transformer framework while still achieving competitive results. The results, summarized in Table \ref{tab:accuracy-comparison}, demonstrate the model's effectiveness compared to both bioplausible and non-bioplausible architectures.

\subsubsection{Sentiment Classification on IMDB}
On the IMDB dataset, the Astromorphic Transformer achieves an accuracy of 88.7\%, outperforming several traditional models, including RNNs, LSTMs, and SNNs. It also exceeds the accuracy of our implementation of Kozachkov \textit{et al}.'s proposal \cite{kozachkov2022building}: Linearized Transformer inspired from astrocytes by 0.3\%, which lacks the astrocytic non-linearity and relative positional encoding central to our model's design. 

\subsubsection{Image Classification on CIFAR10}
In the CIFAR10 dataset, the Astromorphic Transformer attains an accuracy of 97.0\%, outperforming the Linearized Transformer inspired from astrocytes \cite{kozachkov2022building} which achieves 96.8\%. Both models deliver high performance, but the Astromorphic Transformer shows a slight edge, benefiting from enhanced feature learning enabled by astrocytic non-linearity and positional encoding, especially in complex image classification tasks.

\subsubsection{Language Modeling on Wikitext-2}
For language modeling, the Astromorphic Transformer demonstrates superior performance with a perplexity of 33.8 on the Wikitext-2 dataset. In contrast, \textit{Linearized Transformer inspired from astrocytes \cite{kozachkov2022building} fails to converge due to gradient explosion}, yielding a perplexity of 123.3. This stark difference highlights the necessity of incorporating astrocytic non-linearity for stabilizing training and improving model generalization in sequential tasks.

\begin{figure} [t]
\centering
\includegraphics[width=1\columnwidth]{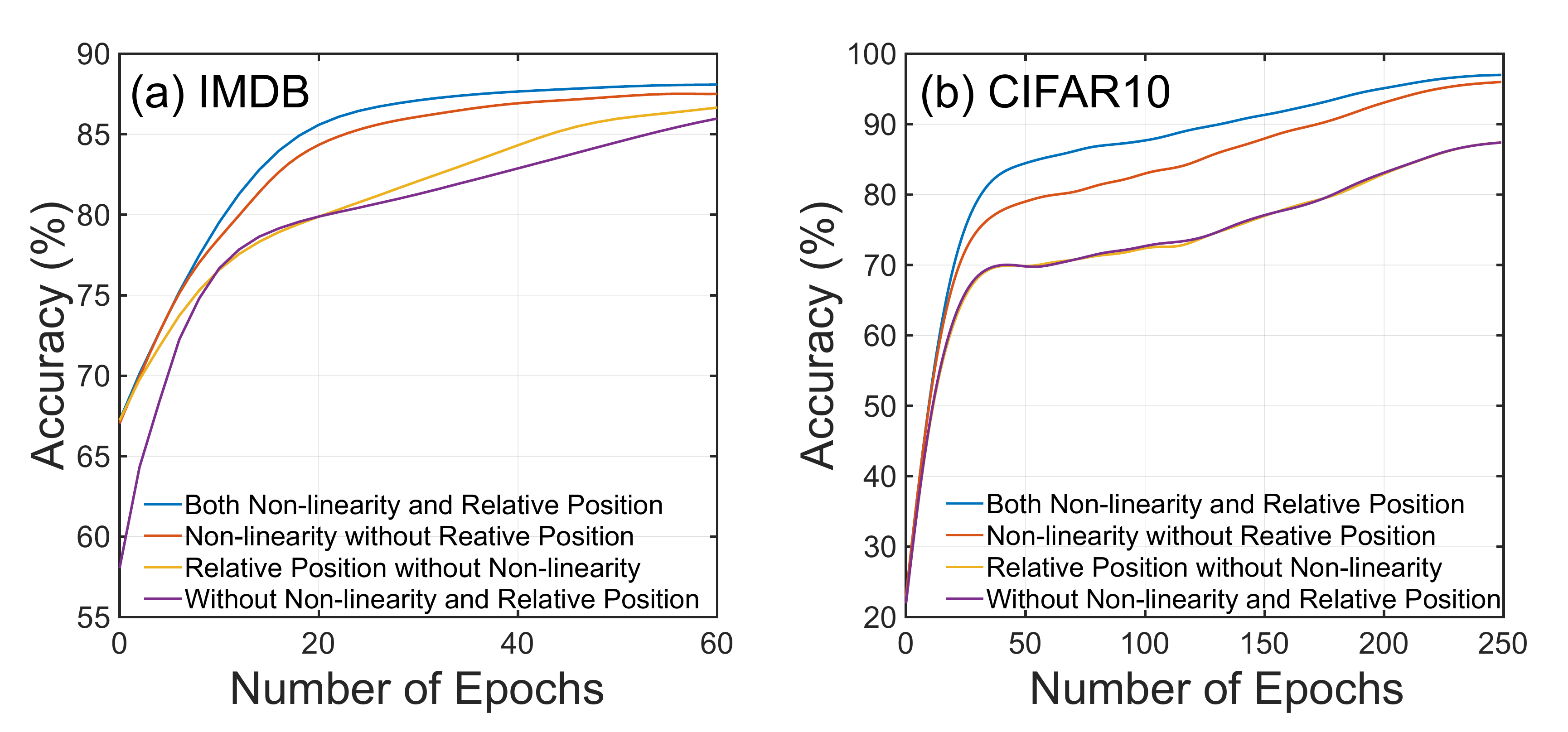} 
\caption{Comparative analysis to examine the effects of incorporating non-linearity and relative positional encoding in the Astromorphic Transformer on the (a) IMDB and (b) CIFAR10 datasets. The yellow and purple lines in Fig. (b) are nearly overlapping. The purple curve represents the result of the model formulated by \cite{kozachkov2022building} which performs subpar in both datasets compared to our proposed Astromorphic Transformer (blue curve). 
}
\label{fig:comparison}
\end{figure}

\subsubsection{Ablation Studies in Learning Speed}
The impact of \textit{astrocytic non-linearity} and \textit{relative positional encoding} on learning speed is evident in both datasets. On the IMDB dataset, the Astromorphic Transformer reaches 85\% accuracy in \textbf{16$\pm$2 epochs}. Without \textit{positional encoding}, this increases to \textbf{22$\pm$2 epochs}, and without both features, it takes \textbf{54$\pm$4 epochs}. These results reflect a \textbf{27.3\% faster convergence} with positional encoding and a \textbf{70.4\% improvement} when both features are included. 
In the CIFAR10 dataset, the model reaches 85\% accuracy in \textbf{45$\pm$3 epochs}. Without \textit{positional encoding}, it requires \textbf{111$\pm$4 epochs}, and without either feature, it stretches to \textbf{215$\pm$2 epochs}, demonstrating a \textbf{59.5\% faster convergence} with positional encoding and a \textbf{79.1\% speed improvement} with both features. For Wikitext-2 dataset, the Astromorphic Transformer converges to the lowest perplexity within $40\pm5$ epochs, whereas the Linearized Transformer inspired from astrocytes \cite{kozachkov2022building} fails to converge, and the model crashes due to gradient explosion after a few epochs. 
Fig. \ref{fig:comparison} makes it evident that our model significantly outperforms the Linearized Transformer inspired from astrocytes \cite{kozachkov2022building} in terms of both learning speed and final performance, demonstrating the crucial role of bioplausible features in improving the convergence and generalization of transformer models.

\section{Conclusion}

The Astromorphic Transformer advances the frontier of bioplausible computing by effectively leveraging neuron-astrocyte interactions to enhance transformer architectures. Through the integration of biologically inspired mechanisms, our model not only matches but also surpasses the efficiency and stability of existing artificial transformer models. By refining synaptic modulation with astrocytic feedback and Hebbian plasticity, we introduce a more adaptive and nuanced attention mechanism that proves advantageous in tasks such as sentiment analysis, image classification, and natural language generation.

While previous models, like the Linearized Transformer inspired from astrocytes \cite{kozachkov2022building}, also utilize neuron-astrocyte interactions, they lack two critical features: astrocytic non-linearity and relative positional encoding. These omissions limit the model’s ability to capture complex relationships in sequential data, resulting in slower convergence and reduced accuracy and, in some cases, failure to converge entirely—as observed in the Wikitext-2 task. Table \ref{tab:transformer_comparison} compares the vanilla transformer, linear transformer, linearized transformer inspired by astrocytes, and the proposed Astromorphic Transformer. It maps each model to relevant equations and figures, emphasizing how computational neuroscience equations underpin the bioplausible mechanisms and advancements of our approach.

\begin{table}[h!]
\centering
\caption{Comparison of Transformer Architectures}
\begin{tabular}{p{1.7cm}||p{6.3cm}}
\hline

Model & Description \\ \hline

Vanilla Transformer & 

Lacks all bioplausibility, employs softmax attention instead of feature maps, and has quadratic complexity for self-attention computation. \\ \hline

Linear Transformer \cite{katharopoulos2020transformers} & 

Lacks bioplausibility, employs linear feature maps (replacing softmax), achieving linearized complexity for self-attention (see Fig. \ref{fig4:order-of-complexity}). Does not incorporate tripartite synapse-based formulation. \\ \hline

Linearized Transformer Inspired from Astrocytes \cite{kozachkov2022building} & 

Incorporates bioplausibility and tripartite synapse-based formulation (refer to Eqns. \ref{eq:write-mode}, \ref{eq:H-neuron}, \ref{eq:PSP}, \ref{eq:cal-resp}, \ref{eq:read-mode} and Fig. \ref{fig2:tripartite synapse}). Employs linear feature maps (no softmax) and has linearized complexity (Fig. \ref{fig4:order-of-complexity}). However, it lacks intricate bioplausible plasticities derived from computational neuroscience models. \\ \hline

Astromorphic Transformer (Our Model) & 

Fully bioplausible, employs linear feature maps (no softmax) with linear complexity (Fig. \ref{fig4:order-of-complexity}) and tripartite synapse formulation (Fig. \ref{fig2:tripartite synapse}). Shows a clear depiction of the astromorphic self-attention network (Fig. \ref{fig3:neural-network}) and incorporates astrocytic non-linearity derived from calcium response (refer to Eqns. \ref{eq:ca-diff-eqn}, \ref{eq:nonlinearity}, \ref{eq:P-weight}, and Fig. \ref{fig5:ca-response}). Introduces astrocytic Hebbian plasticities (Eqns. \ref{eq:H-astro}, \ref{eq:HP}) encoding relative positional information (Eqn. \ref{eq:edge-info}) inspired by Slow Inward Currents (Eqns. \ref{eq:mem-potential}, \ref{eq:synaptic-current}). Formulates an accurate model of the Astromorphic Transformer, as detailed in Eqn. \ref{eq:astro-transformer-matrix}. \\ \hline
\end{tabular}
\label{tab:transformer_comparison}
\end{table}

Moreover, our Astromorphic Transformer is catered for simple implementation in current hardware without any additional overhead by leveraging the decomposition of astromorphic attention into distinct write and read modes, making it feasible for traditional computing architectures. Detailed molecular-level computational modeling is not required for practical implementation. Instead, macro-models developed for non-linearity, plasticity, and feedback processes are being utilized, which represent these dynamics in a simplified yet biologically plausible way. These macro-models (see Equation \ref{eq:astro-transformer-matrix}) should be easily integrable into conventional hardware, facilitating efficient deployment without needing highly specialized configurations.
Energy-efficient in-memory processing with CMOS or emerging post-CMOS devices for transformer architectures \cite{yang2020retransformer} can also be potentially utilized for hardware implementation of Astromorphic Transformers. 
The Astromorphic Transformer can further be improved by incorporating more complex dynamics from neuroscience, such as astrocyte-astrocyte communication, which could enable more sophisticated processing and parallelization of plasticity mechanisms. Additionally, scaling the model to deeper architectures while leveraging characteristics like memory and learning from astrocytes may allow the model to adapt to more complex tasks with long-context sequences.

\section*{Acknowledgments}

Research was sponsored in part by the Army Research Office and was accomplished under Grant Number W911NF-24-1-0127. The views and conclusions contained in this document are those of the authors and should not be interpreted as representing the official policies, either expressed or implied, of the Army Research Office or the U.S. Government. The U.S. Government is authorized to reproduce and distribute reprints for Government purposes notwithstanding any copyright notation herein. This work was also supported partially by the National Science Foundation under award No. EFRI BRAID \#2318101.

\clearpage

\pagebreak
\renewcommand{\thefigure}{S\arabic{figure}}
\renewcommand{\thetable}{S\arabic{table}}
\setcounter{figure}{0}   
\setcounter{table}{0}   
%\beginsupplement
%\begin{document}
\section*{Supplementary Material}
\subsection{Appendix A: Transformer Illustration}
\begin{figure} [h!]
\centering
\includegraphics[width=0.5\columnwidth]{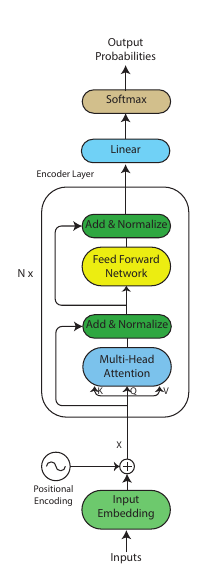} 
\caption{Schematic representation of the Transformer Encoder architecture, illustrating the sequential flow of operations. $N$ denotes the number of encoder layers in the network.}
\label{fig1:transformer}
\end{figure}

Transformer employs only encoder architecture for performing sentiment classification and vision tasks. As illustrated in Fig. \ref{fig1:transformer}, the input embeddings, combined with the positional embeddings, enter the encoder layer as input tokens. Then \textit{key}, \textit{query}, and \textit{value} are generated from the input tokens, and after performing self-attention on them, a skip connection is added from the inputs, and the overall result is layer normalized and fed into a feed-forward network. Another add \& normalize operation is performed before feeding them into a linear layer and softmax function to obtain the output probabilities. Sometimes, there might be more than one encoder layer. For different tasks, like machine translation and text generation, encoder-decoder or only decoder-type architecture of transformers can be used. However, independent of the type of architecture needed, self-attention is a fundamental mechanism defining transformer operation. The ``Multi-Head Attention" Block has self-attention modules parallelly running on different heads obtained from the input embeddings.

\subsection{Appendix B: Memory Efficient Linearized Attention}
\label{linearized-attention}
\begin{figure}[h!]
\centering
\includegraphics[width=0.85\columnwidth]{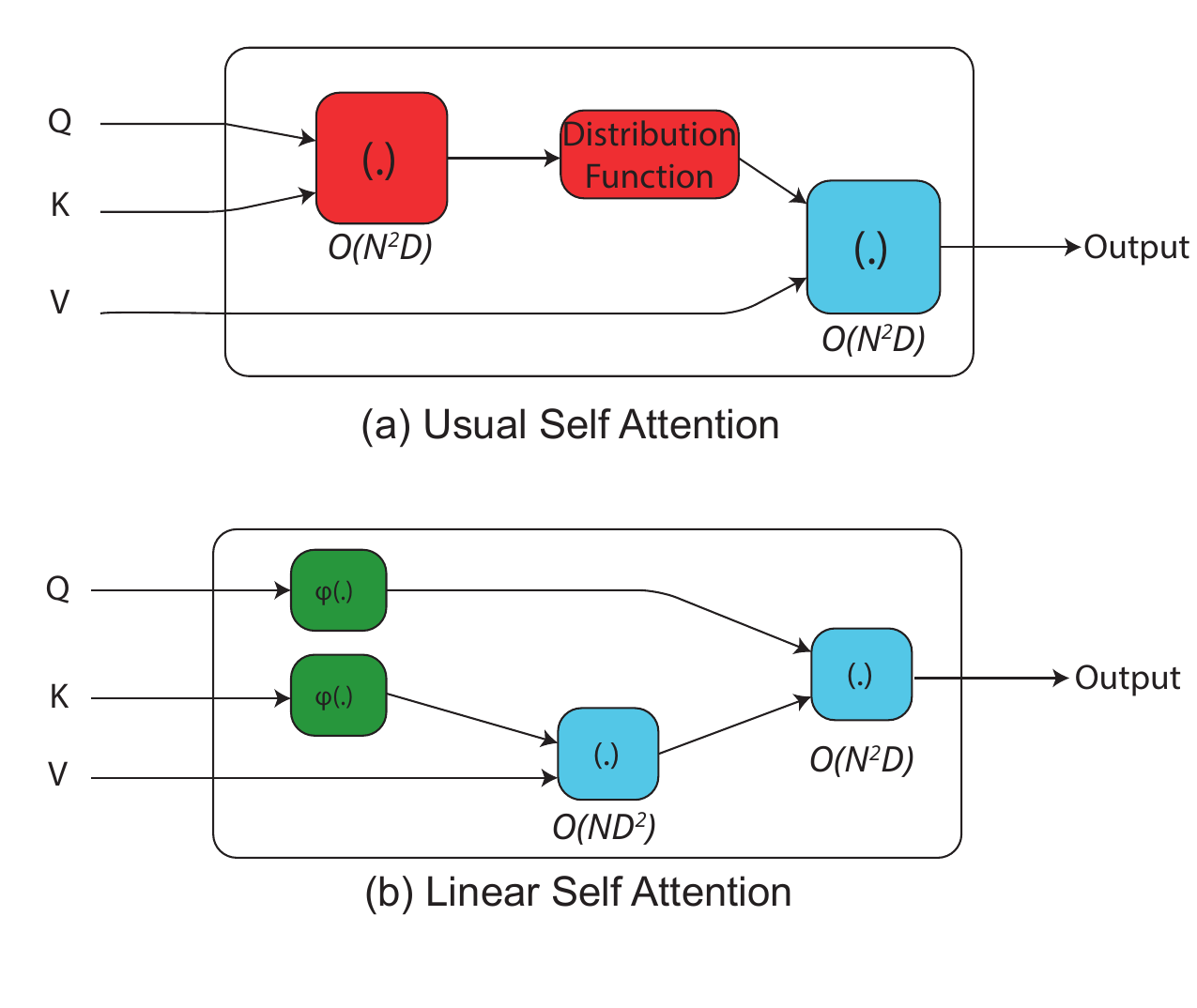} 
\caption{Figure showing the comparison of order of complexity between usual and linearized self-attention. Distribution function can be mapped to softmax function used in usual transformers.}
\label{fig4:order-of-complexity}
\end{figure}

Linearized attention involves utilizing feature maps with lower complexity levels, thereby minimizing memory and computational resources. As discussed before, the introduction of non-linearity in biological neurons is facilitated by an activation function ($\phi(.)=elu(.)+1$) in the hidden layer neurons. The utilization of this positive similarity feature map as an activation function characterizes the implementation of the linearized attention operation as described by Equation 1. The utilization of linearized feature attention results in a reduction of computational complexity associated with the computation of the attention score. The advantage of utilizing feature map-based linearized attention over the conventional distribution function (\textit{i.e.} softmax) based self-attention is demonstrated in Fig. \ref{fig4:order-of-complexity}.
Usual self-attention employs softmax as the distribution function; however, the dot product of \textit{Query ($Q$)} and \textit{Key ($K$)} poses an \textit{$O(N^2D)$} complexity, which overuses the memory. Whereas, in linearized attention, the computation multiplies \textit{Key ($K$)} and \textit{Value ($V$)}, yielding an \textit{$O(ND^2)$} complexity, properly utilizing the memory and computational efficiency when token length ($N$) is larger than internal embedding size ($D$)(We set token embedding dimension equal to the internal embedding size ($d=D$) in this work). Thus, the employment of the linearized feature map enables the realization of memory-efficient linearized attention.

\subsection{Appendix C: Hyper-parameter Details} \label{sup:hyp}
Hyper-parameters for the encoder-type astromorphic transformer for IMDB dataset are in Table \ref{tab:hyp_IMDB}, for the encoder-type convolution and astromorphic self-attention-based transformer for CIFAR10 dataset are in Table \ref{tab:hyp_CIFAR10}, and for the astromorphic transformer for Wikitext-2 dataset are in Table \ref{tab:hyp_Wikitext2}.
\begin{table}
    \small
    \caption{Hyper-parameter list for the IMDB dataset.}
    \centering

    \begin{tabular}{ m{12em}|| m{4em} | m{6em} }  %{|c|c|c|c|c|c|} %

     \hline

     Hyper-parameters & Value & Tested Range\\ 

     \hline

     \hline

     Non-linearity parameter $\alpha$ & $0.25$ & $0.15$-$0.4$\\

     \hline

     Model size $D$ & $128$ & $32$-$256$\\  

     \hline

     Scaling parameter $m$ & $500D$ & ($1$-$1000$)$D$\\     

     \hline

     Sequence length  & $400$ & $100$-$600$\\  

     \hline

     Learning rate & $0.00001$ & $1e^{-4}$ - $1e^{-6}$\\  

     \hline

     Batch size & $64$ & $8$-$256$\\  

     \hline

     Number of epochs & $120$ & $40$-$150$\\  

     \hline

     Number of heads & $4$ & $1$-$12$\\  

     \hline
     \end{tabular}
    \label{tab:hyp_IMDB}
\end{table}

\begin{table}
    \small
    \caption{Hyper-parameter list for the CIFAR10 dataset.}
    \centering

    \begin{tabular}{ m{12em}|| m{4em} | m{6em} }  %{|c|c|c|c|c|c|} %

     \hline

     Hyper-parameters & Value & Tested Range\\  

     \hline

     \hline

     Non-linearity parameter $\alpha$ & $0.25$& $0.15$-$0.4$\\ 

     \hline

     Model size $D$ & $256$ & $64$-$512$\\  

     \hline

     Scaling parameter $m$ &  $500D$ & ($1$-$1000$)$D$\\

     \hline

     Patch size & $2$ & $2$-$4$\\  

     \hline

     Learning rate & $0.001$& $1e^{-2}$ - $1e^{-4}$\\  

     \hline

     Batch size & $32$ & $8$-$128$\\  

     \hline

     Number of epochs & $250$& $80$-$300$\\  

     \hline

     Number of heads & $8$& $2$-$12$\\  

     \hline

     \end{tabular}
    \label{tab:hyp_CIFAR10}
\end{table}

\begin{table}[t]
    \small
    \caption{Hyper-parameter list for the Wikitext-2 dataset.}
    \centering

    \begin{tabular}{ m{12em}|| m{4em} | m{6em} }  %{|c|c|c|c|c|c|} %

     \hline
     Hyper-parameters & Value & Tested Range\\  

     \hline

     \hline

     Non-linearity parameter $\alpha$ & $0.25$& $0.1$-$0.3$\\ 

     \hline

     Model size $D$ & $768$ & $256$-$1024$\\  

     \hline

     Scaling parameter $m$ &  $100D$ & ($50$-$500$)$D$\\

     \hline

     Sequence length & $512$ & $128$-$512$\\  

     \hline

     Learning rate & $0.00001$& $1e^{-4}$ - $5e^{-6}$\\  

     \hline

     Batch size & $24$ & $8$-$64$\\  

     \hline

     Number of epochs & $50$& $20$-$100$\\  

     \hline

     Number of heads & $6$& $4$-$12$\\  

     \hline

     \end{tabular}
    \label{tab:hyp_Wikitext2}
\end{table}
\subsection{Appendix D: Computational Model of Neuron-Astrocyte-Mediated Synaptic Dynamics }
In the context of astromorphic computation, astrocytes play a pivotal role in modulating synaptic transmission through both direct and indirect pathways. These pathways, influenced by endocannabinoid signaling and astrocytic calcium dynamics, are essential for the dynamic control of synaptic weights in neuromorphic architectures such as the Astromorphic Transformer.

\subsubsection{Endocannabinoid Dynamics}
Endocannabinoids, such as 2-arachidonoylglycerol (2-AG), are released from the postsynaptic neuron following depolarization. They serve as retrograde messengers, modulating synaptic transmission probability through both depressive and potentiating effects. The release dynamics of 2-AG can be modeled as follows:
\begin{equation*}
    \frac{d(AG)}{dt} = -\frac{AG}{\tau_{AG}} + r_{AG} \cdot \delta(t - t_{\text{spike}})
\end{equation*}
where \( \tau_{AG} \) is the decay constant of 2-AG, and \( r_{AG} \) is the rate of 2-AG production after a spike.

\subsubsection{Direct Signaling Pathway—Depolarization-Induced Suppression of Excitation (DSE)}
In the direct pathway, 2-AG binds to CB1 receptors (CB1Rs) on the presynaptic neuron, reducing the synaptic transmission probability \( PR \). This localized effect, termed Depolarization-Induced Suppression of Excitation (DSE), can be described as:
\begin{equation*}
    DSE(t) = -AG(t) \cdot K_{AG}
\end{equation*}
where \( K_{AG} \) is a constant representing the strength of suppression.
\subsubsection{Indirect Signaling Pathway—Endocannabinoid-Mediated Synaptic Potentiation (e-SP)}
In the indirect pathway, 2-AG binds to CB1Rs on astrocytes, triggering intracellular calcium release from the endoplasmic reticulum (ER) via IP3 receptors (IP3Rs). This calcium release induces astrocytic glutamate release, which potentiates synaptic transmission at distant synapses. The potentiation is given by:
\begin{equation*}
    eSP(t) = m_{eSP} \cdot Glu(t)
\end{equation*}
where \( m_{eSP} \) is the scaling factor for synaptic potentiation, and \( Glu(t) \) is the concentration of astrocytic glutamate.

\subsubsection{Calcium Dynamics in Astrocytes}
The astrocytic calcium response, which regulates synaptic potentiation, is governed by the following equation:
\begin{equation*}
    \frac{d[Ca^{2+}]}{dt} = J_{\text{chan}}(IP3, Ca^{2+}) + J_{\text{leak}} - J_{\text{pump}}
\end{equation*}
where \( J_{\text{chan}}(IP3, Ca^{2+}) \) represents calcium influx through IP3-regulated channels (dependent on IP3 and calcium concentration), \( J_{\text{leak}} \) accounts for passive calcium leakage from the ER, and \( J_{\text{pump}} \) describes calcium reuptake into the ER via ATP-dependent pumps.
Once calcium levels exceed a threshold, the astrocyte releases glutamate, governed by:
\begin{equation*}
    \frac{d[Glu]}{dt} = -\frac{Glu}{\tau_{Glu}} + r_{Glu} \cdot H(Ca^{2+}(t) - \theta_{Ca})
\end{equation*}
where \( \tau_{Glu} \) is the decay constant of glutamate, and \( \theta_{Ca} \) is the calcium threshold that triggers release, and \( H(x) \) is the Heaviside function defined as \( H(x) = 1 \) if \( x \geq 0 \) and \( H(x) = 0 \) if \( x < 0 \).

\subsubsection{Modulation of Synaptic Transmission Probability (PR)}
The total synaptic transmission probability \( PR(t) \) is influenced by both the direct depressive effect and indirect potentiation, as follows:
\begin{equation*}
    PR(t) = PR(0) + PR(0) \cdot \frac{DSE(t) + eSP(t)}{100}
\end{equation*}
where \( PR(0) \) is the baseline synaptic transmission probability.
%\subsubsection{Relevance to Astromorphic Transformers}
%This model of neuron-astrocyte interactions provides a foundation for understanding calcium-mediated modulation of synaptic weights in the Astromorphic Transformer. The balance between synaptic depression (DSE) and potentiation (e-SP) allows for dynamic control of synaptic transmission probabilities, which are essential for adaptive and bio-plausible self-attention mechanisms in such neuromorphic architectures.

This appendix presents a detailed mathematical framework for neuron-astrocyte-mediated synaptic dynamics, which are key to the functionality of astromorphic computational systems. %By leveraging astrocytic calcium signaling and endocannabinoid-mediated pathways, synaptic weights can be dynamically adjusted, mimicking the flexible nature of biological networks.

\subsection{Appendix E: Algorithm and Pseudo-code} \label{sup:pseudo-code}
The pseudo-code (Algorithm \ref{alg:pseudo-code}) represents the core steps of the astromorphic transformer, where neuron-astrocyte interactions modulate synaptic weights, enhancing the self-attention mechanism. The algorithm operates through two phases: Write Mode and Read Mode, followed by the generation of the final transformer output.
In the initialization step, the weight matrices for neurons and astrocytes, the presynaptic plasticity parameter, the non-linearity parameter, and the edge matrix representing token positions are set up. During the Write Mode, keys and values are generated from the input tokens using learnable weight matrices. The activation function is applied to introduce non-linearity, resulting in an updated representation. Hebbian learning is then used to update both neuron and astrocyte weight matrices based on the outer product of the activation and the values, while the presynaptic plasticity parameter is updated according to the sum of activated keys modulated by a non-linearity parameter.
In the Read Mode, after all tokens have been processed, a query matrix is generated from the input tokens. The activation function is applied to these queries, and the astrocytic calcium response is computed. The modulation factor, which is inversely proportional to the calcium response, adjusts the synaptic weights dynamically. The final output is calculated by matrix multiplication of the synaptic weights and the activated queries along with an addition of a residual connection from the input.
The final transformer output is processed through layer normalization to stabilize and accelerate training. A feed-forward network introduces additional non-linearity, followed by another layer normalization step with a residual connection to ensure gradient flow. Finally, a linear transformation and softmax function produce the final logits, representing the output probabilities of the model.

\begin{algorithm} 
\caption{Astromorphic Transformer Algorithm}
\begin{algorithmic}[1] \label{alg:pseudo-code}
    \STATE \textbf{Initialization:} Initialize parameters and hyper-parameters
    \STATE Initialize input tokens $X$, token embedding dimension $d$, hidden layer neurons $m$
    \STATE Initialize Hebbian weight matrix $H_{neuron}$ and astrocyte weight matrix $H_{astro}$ to 0
    \STATE Initialize presynaptic plasticity parameter $g$ to 0
    \STATE Set non-linearity parameter, $\alpha$
    \STATE Set edge between input tokens $x_i$ and $x_j$: $a_{ij}$    
    \STATE \textbf{Write Mode:} Encode keys and values from input tokens
    \FOR{each token $x_t$ in input tokens $X$}
        \STATE $k_t \leftarrow x_t  W_{k}$  %\COMMENT{Generate key using weight matrix}
        \STATE $v_t \leftarrow x_t  W_{v}$  %\COMMENT{Generate value using weight matrix}
        \STATE $h_t \leftarrow \phi(k_t)$  %\COMMENT{Apply activation for non-linearity}
        \STATE \COMMENT{Hebbian learning for neuron and astrocyte plasticity}
        \STATE $H_{neuron,t} \leftarrow H_{neuron,0} + \frac{1}{m} h_t^T  v_t$
        \STATE $W_{astro} \leftarrow \phi(a_{ij}^T)$ 
        \STATE $H_{astro,t} \leftarrow H_{astro,0} + \frac{1}{m} W_{astro}  v_t$
        \STATE \COMMENT{Presynaptic plasticity based on calcium dynamics}
        \STATE $g \leftarrow g + (\sum(h_t))^\alpha$
    \ENDFOR
    \STATE \textbf{Read Mode:} Retrieve information using query for all tokens
    %\FOR{each query $q_t$ in queries, $Q$}
    \STATE $Q \leftarrow X  W_{q}$  %\COMMENT{Generate query using weight matrix}
    \STATE $h \leftarrow \phi(Q)$  %\COMMENT{Apply activation for non-linearity}
    \STATE $C \leftarrow h  g^T$  %\COMMENT{Calcium response from astrocyte}    
    %\STATE \COMMENT{Calculate weight $P$ based on astrocyte modulation (inverse relation with calcium response)}
    \STATE $P \leftarrow \frac{1}{C}$
    %\STATE \COMMENT{Final output using Hebbian and astrocyte weights, modulated by $P$}
    \STATE $L \leftarrow h  ((H_{neuron} + H_{astro}) \odot P) + X$
%\ENDFOR
    %\STATE \COMMENT{Segmented computation for the final Transformer output}
    \STATE $L_{norm} \leftarrow \mathrm{LayerNorm}(L)$
    \STATE $L_{output} \leftarrow \mathrm{LayerNorm}(\mathrm{FFN}(L_{norm}) + L_{norm})$  %\COMMENT{Layer normalization with residual input connection}
    \STATE $logits \leftarrow \mathrm{Softmax}(\mathrm{Linear}(L_{output}))$ %\COMMENT{Transformer output probabilities}
    \RETURN logits
\end{algorithmic}
\end{algorithm}
\color{black}

\end{document}